\documentclass[letterpaper, 10 pt, conference]{ieeeconf}  % Comment this line out if you need a4paper
\usepackage{graphicx} % Required for inserting images
\usepackage{amsmath,amsfonts}
\usepackage{algorithmic}
\usepackage{algorithm}
\usepackage{array}
\usepackage[caption=false,font=normalsize,labelfont=sf,textfont=sf]{subfig}
\usepackage{textcomp}
\usepackage{stfloats}
\usepackage{url}
\usepackage{verbatim}
\usepackage{graphicx}
\usepackage{svg}
\usepackage{cite}
\usepackage{float}
\usepackage{balance}
\usepackage[export]{adjustbox}
\usepackage[colorlinks=true,
            citecolor=blue,
            linkcolor=red,
            urlcolor=cyan
            ]{hyperref}

\IEEEoverridecommandlockouts % Fixes the \thanks not working

\title{\LARGE \bf
Manufacturing Micro-Patterned Surfaces with Multi-Robot Systems
}

\author{Annalisa T. Taylor, Malachi Landis, Ping Guo, and Todd D. Murphey% <-this % stops a space
\thanks{Mechanical Engineering Department, Northwestern University}%
\thanks{Corresponding authors: Ping Guo \href{mailto:ping.guo@northwestern.edu}{ping.guo@northwestern.edu} and Todd Murphey \href{mailto:t-murphey@northwestern.edu}{t-murphey@northwestern.edu} }%
}

\begin{document}

\maketitle
\thispagestyle{empty}
\pagestyle{empty}

\begin{abstract}
Applying micro-patterns to surfaces has been shown to impart useful physical properties such as drag reduction and hydrophobicity. However, current manufacturing techniques cannot produce micro-patterned surfaces at scale due to high-cost machinery and inefficient coverage techniques such as raster-scanning. In this work, we use multiple robots, each equipped with a patterning tool, to manufacture these surfaces. To allow these robots to coordinate during the patterning task, we use the ergodic control algorithm, which specifies coverage objectives using distributions. We demonstrate that robots can divide complicated coverage objectives by communicating compressed representations of their trajectory history both in simulations and experimental trials. Further, we show that robot-produced patterning can lower the coefficient of friction of metallic surfaces. This work demonstrates that distributed multi-robot systems can coordinate to manufacture products that were previously unrealizable at scale.
\end{abstract}

% \thanks{This work was supported by the National Science Foundation under grant CNS-2229170.}% <-this % stops a space

\section{Introduction}

Micro-patterned surfaces have been shown to offer passive physical benefits, including friction control, drag reduction, and anti-fouling properties~\cite{yeung2019friction, pawelski1994influence}. These properties could be advantageous when applied to large workpieces, such as the hulls of ships, to improve fuel efficiency. However, existing manufacturing techniques struggle to produce such surfaces at scale without high-precision equipment. Traditional motion planning strategies for manufacturing, such as raster scanning, require micro-scale features to be placed over large areas in predetermined locations. As a result, no scalable method exists for manufacturing micro-patterned surfaces. 

\begin{figure}[ht]
\centering
\includegraphics[width=0.99\columnwidth]{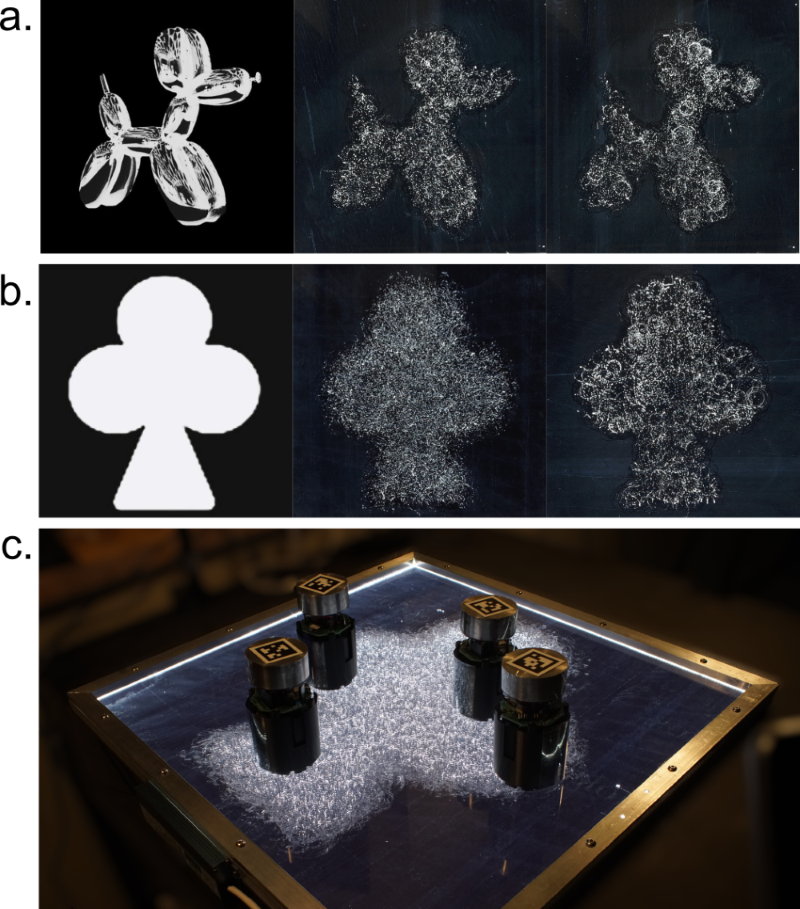}
    \caption{\textbf{Robot patterning.}~(a) Four robots patterning an image of a sculpture in the \textit{Balloon Dog} series by Jeff Koons. From left to right: coverage objective, patterning by four robots without communication, patterning by four communicating robots. Communicating robots use the decentralized ergodic control algorithm, in which agents average their trajectory history to divide coverage. (b) From left to right: the club coverage objective, patterning by four non-communicating robots, patterning by robots sharing their trajectory history. (c) Robots patterning the club objective on acrylic.}
    \label{fig:fig_1_multi}
    \vspace{-2em}
\end{figure}

\begin{figure}[t!]
\centering
\includegraphics[width=0.95\columnwidth]{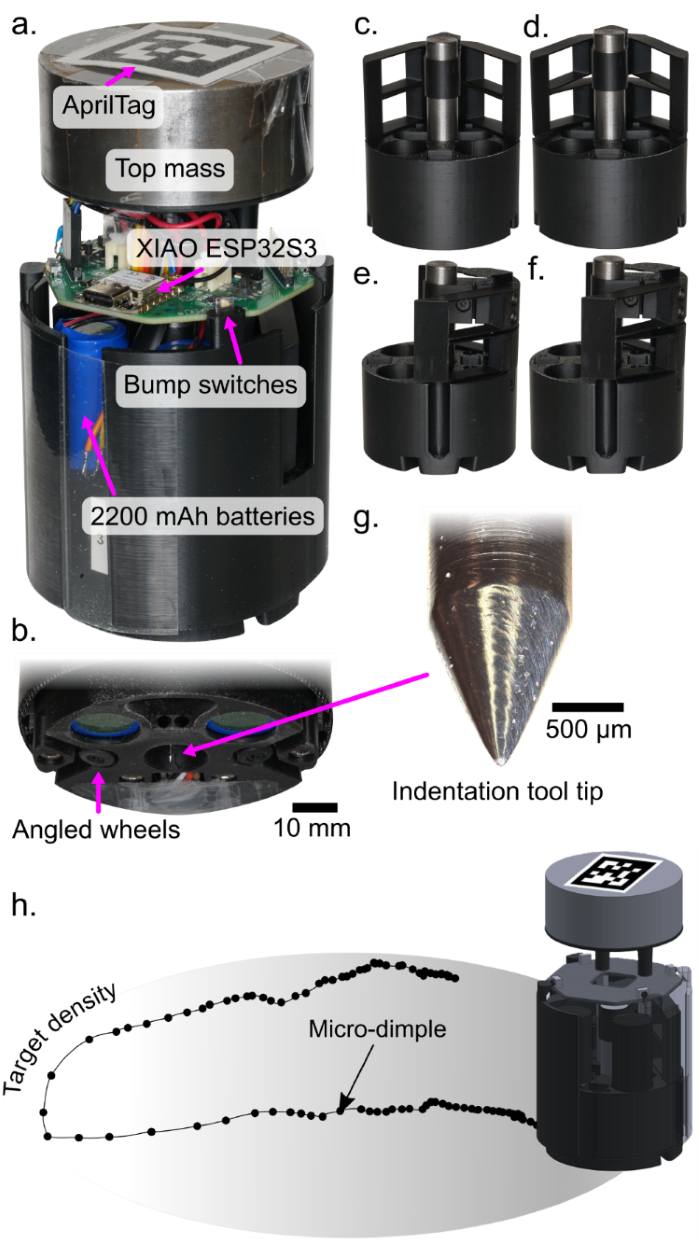}
    \caption{\textbf{Robot design.} The robot (a) uses an ESP32S3 microcontroller on a custom PCB. Angled wheels (b) minimize the footprint. The indentation tool moves from a low (c,e) to a high (d,f) position using a cam, which then releases and allows the tip (g) to impact the workpiece, forming a micro-dimple. (h) The robot moves according to the target density distribution to place these micro-dimples. }
    \label{fig:robot_diagram}
\end{figure}

In this work, we show that these requirements are not necessary to create useful micro-patterns, as the underlying physical properties that enable surface functionality depend on feature density, not exact feature placement. This presents an opportunity to rethink the manufacturing process using the principles underlying ergodic control, where the objective is not precise localization, but time allocation proportional to a spatial distribution~\cite{miller2016exploration, taylor2021active}. To this end, we use mobile robots with an indentation tool onboard, which apply micro-dimples to a workpiece to micro-pattern surfaces, as shown in Figure~\ref{fig:robot_diagram}. Previously, the foundation for scaling these techniques to collaborative teams was established using a single robot for micro-patterning~\cite{TaylorIROS2024}. This robot used an indentation tool to apply divots to a metallic surface through plastic deformation following pre-planned trajectories generated with the ergodic control algorithm, developed in~\cite{mathew2011metrics, miller2016ergodic}. 
Subsequent work showed that robots computing ergodic trajectories could pattern independently on the same metallic workpieces~\cite{landis2025patterning}. Here, we extend that work by allowing agents to share their trajectory histories and show that patterning produced by mobile robots can impart beneficial physical properties to metallic surfaces. This strategy allows these compact, mobile robots to generate micro-scale patterns that match given target density distributions, consistent with desired physical effects, such as drag reduction. We verify the density-based specification and robot performance in hardware by evaluating a robot-produced patterning in a lubricated friction test. We find that the robots produce the desired patterns and achieve a reduction in sliding friction on a metallic workpiece.~\textbf {The primary contributions of this work are listed below:}
\begin{enumerate}
    \item We develop a manufacturing technique for scalable, multi-robot surface micro-patterning using density specifications and demonstrate this process in simulated and hardware experiments.
    \item We test the tribological properties of robot-patterned surfaces and show that robot patterning can reduce friction on a metallic surface.
    \item We demonstrate that communication facilitates task decomposition in simulated and experimental systems.
\end{enumerate} 

These results demonstrate that distributed robotic systems can manufacture micro-patterned surfaces at scale.

\section{Decentralized Ergodic Control}
\label{sec:ergodic_alg}
Traditional manufacturing techniques focus on optimizing precision-based objectives: following tight tolerances to make products in an assembly line as similar as possible without defects. This is often the right instinct for creating reliable products, but achieving high-precision results can limit scalability. However, in the case of surface micro-patterning, existing literature assumes the physical benefits induced by micro-patterned surfaces rely on feature density~\cite {vilhena2009surface, dan2020tribological,tong2021properties}. 

To synthesize robot controls that allow density functions to be used as manufacturing objectives, we use the ergodic metric introduced in~\cite{mathew2011metrics}. In our system, each robot runs an ergodic optimal control algorithm onboard to cover a target density function, spending time in proportion to the value of the target density over the surface~\cite{miller2016exploration, taylor2021active}. This allows robots to traverse a workpiece quickly while still completing the underlying coverage task. Each robot applies dimples to the surface at a constant rate and plans its trajectory to modulate pattern density, as shown in Figure~\ref{fig:fig_1_multi}. Our method, shown in Figure~\ref{fig:robot_diagram}h, enables the robot to pattern target images defined by density objectives while producing functional surfaces. 

Manufacturing errors on large workpieces are often costly. To ensure that agents do not pattern undesired areas of the workpiece, we use control barrier functions to confine an agent to a safe subset of the coverage domain~\cite{ames2019control}. Here, control barrier functions are added as a penalty in the controller cost function. Enforcing safety in this work equates to protecting the manufactured product from robot errors instead of ensuring the safety of the robots themselves.

\subsection{Ergodic Metric}
Ergodicity is a statistical property of dynamical systems in which time is spent in regions of space in proportion to the measure of those regions~\cite{krengel1985ergodic}. This makes ergodic dynamics with respect to desired measures useful for coverage and exploration tasks in robotics~\cite{shellMult05ergodic,prabhakar2020ergodic, salman2017multi,mathew2011metrics}.
We use the ergodic control algorithm, originally developed in~\cite{mathew2011metrics}, to generate control actions that match trajectory statistics to spatial distributions. This enables an ergodic controller to cover a given task distribution without additional metrics for dividing the coverage domain~\cite{miller2016exploration}. In prior work, an ergodic controller was used to generate trajectories offline for a single-agent patterning task~\cite{taylor2024patterning}. In this work, we use the decentralized ergodic control algorithm for coverage of target spatial distributions with a multi-robot system~\cite{abraham2018decen}. To describe an optimal controller for a multi-robot system, we assume the system dynamics of $N$ agents are in control-affine form, controllable, and independent:
\begin{align} \label{eq:collective_dynamics}
    \dot{x} & = f(x) + g(x) u \nonumber\\
    & = \begin{bmatrix}
    f_1(x_1) \\
    \vdots \\
    f_N(x_N)
    \end{bmatrix} +
    \begin{bmatrix}
    g_1(x_1) & \ldots & 0\\
    \vdots& \ddots & \\
    0 & & g_N(x_N)
    \end{bmatrix} u,
\end{align}
where the state of a robotic agent at every point in time $t\in \mathbb{R}^+$ is $x(t) \in \mathbb{R}^n$ and the control is $u(t) \in [u_{min}, u_{max}] \subset \mathbb{R}^m$. Where $f(x): \mathbb{R}^n \to \mathbb{R}^n$ and $g(x): \mathbb{R}^n \to \mathbb{R}^{n \times m}$ are vector fields for the system's free dynamics and control input response. The trajectories of the multi-agent system are represented by: $x(t) = \left[ x_1(t), \ldots, x_N(t)\right]\in \mathbb{R}^{nN}$. We use a receding-horizon optimal control algorithm to generate actions in time for agents over closed subsets of $\mathbb{R}^+$; we have $t \in \left[t_i, t_i+\tau\right]$ for some initial time $t_i$ with the $i^\text{th}$ sampling time and time horizon $\tau>0$. 

We now define a metric on ergodicity, which will be evaluated using spatial Fourier coefficients. The Fourier coefficients for the trajectory statistics $x(t)$ are:
\begin{equation}
\label{eq:trajectory_representation}
    c_k= \frac{1}{\tau} \int_{t_i}^{t_i+\tau} \tilde{F}_k(x(t)) dt.
\end{equation}
Note that $\tilde{F}_k(x(t)) = \frac{1}{N}\sum_j F_k(x_j(t))$ where $j$ is the robot index and $F_k(x)$ are cosine basis functions~\cite{abraham2018decen}. We also require a Fourier representation of the target distribution. Here, this will be the goal patterning distribution, given by $ \phi_k =  \int_{T} \phi(x) F_k(x) dx$, where $\phi(x)$ is the measure of the distribution over the task space. Finally, we write an ergodic metric:
\begin{equation} \label{eq:ergodic_metricCK}
    E(x(t)) = \,\sum_{k \in \mathbb{N}^n} \Lambda_k \left(c_k -\phi_k \right)^2
\end{equation}
where $\Lambda_k$ is a normalization coefficient ensuring convergence of Equation~\ref{eq:ergodic_metricCK}. Note that the trajectory statistics of the collective $c_k$ are the result of averaging individual agents' trajectory statistics, $c_{k_j}$. This is the mechanism that allows them to coordinate their behavior.

% \begin{figure}
% \centering
% \includegraphics[width=\columnwidth]{figs/dimpling_schematic.png}
%     \caption{\textbf{Process schematic and robot design.} figure that shows the robot and what surface patterning does. Robots apply micro-dimples to a surface, planning their trajectories to dimple more in higher valued areas of a given target density. \textcolor{red}{Note to Malachi: We could add more panels to this figure to show the robot design.}}
%     \label{fig:robot_diagram}
% \end{figure}

\section{Robotic System Design}
\label{sec:robot_design}

Unlike conventional manufacturing tools, such as gantries, mobile robots are not limited by domain size (when the workpiece grows, the tool itself does not need to scale accordingly) and offer an alternative to traditional precision manufacturing techniques for surface patterning~\cite{zhao2023}. For this reason, we have developed a mobile robot to apply surface patterns using a dimpling tool, shown in Figure~\ref{fig:robot_diagram}. Previous work introduced the design of the robot platform in detail, but it will be described briefly here~\cite{landis2025patterning}. 
% Engineering drawings and a bill of materials can be found in our repository~\cite{landis2025github}. 
The robot uses two wheels, angled to minimize the footprint, to traverse flat workpieces. It applies indentations using a sharpened carbide tip which is raised and snapped downwards by a cam-flexure mechanism. Controls are computed onboard using a XIAO ESP32S3 microcontroller powered by two 2200 mAh batteries for at least 2 hours of runtime. Collisions with other robots are detected by bump switches. The majority of the robot, including the flexure mechanism, is FDM 3D printed. The robot itself is approximately 75 mm in diameter and 135 mm tall, small enough to allow multiple robots to work together in a 500 by 500 mm domain. 

\section{Multi-Robot Communication}
\label{sec:robot_comm}
In this work, we test simulated and experimental systems of four robots with two options for communication. In both cases, each robot computes optimal ergodic controls online. First, agents do not explicitly communicate. Each agent is responsible for the same patterning objective, but no information is passed between individuals. Second, robots communicate explicitly by sharing the Fourier representation of their trajectory statistics, as shown in Equation~\ref{eq:trajectory_representation}. Then, robots average the trajectory statistics of all agents and generate their control action based on the history of the collective. This averaging process allows robots to track the patterning progress of the whole system and focus on particular subsections of the coverage domain, which reduces redundant coverage. We refer to the parallelization of the coverage task between agents as task division. To measure task division, we will quantify the difference between each agent's trajectory. Agents that are not communicating are individually responsible for the entire patterning task and thus should have similar trajectories. Agents that have knowledge of the state of the collective patterning process should be able to divide the task, meaning their trajectories will be different from one another.

\begin{figure}[t!]
\centering
\includegraphics[width=\columnwidth]{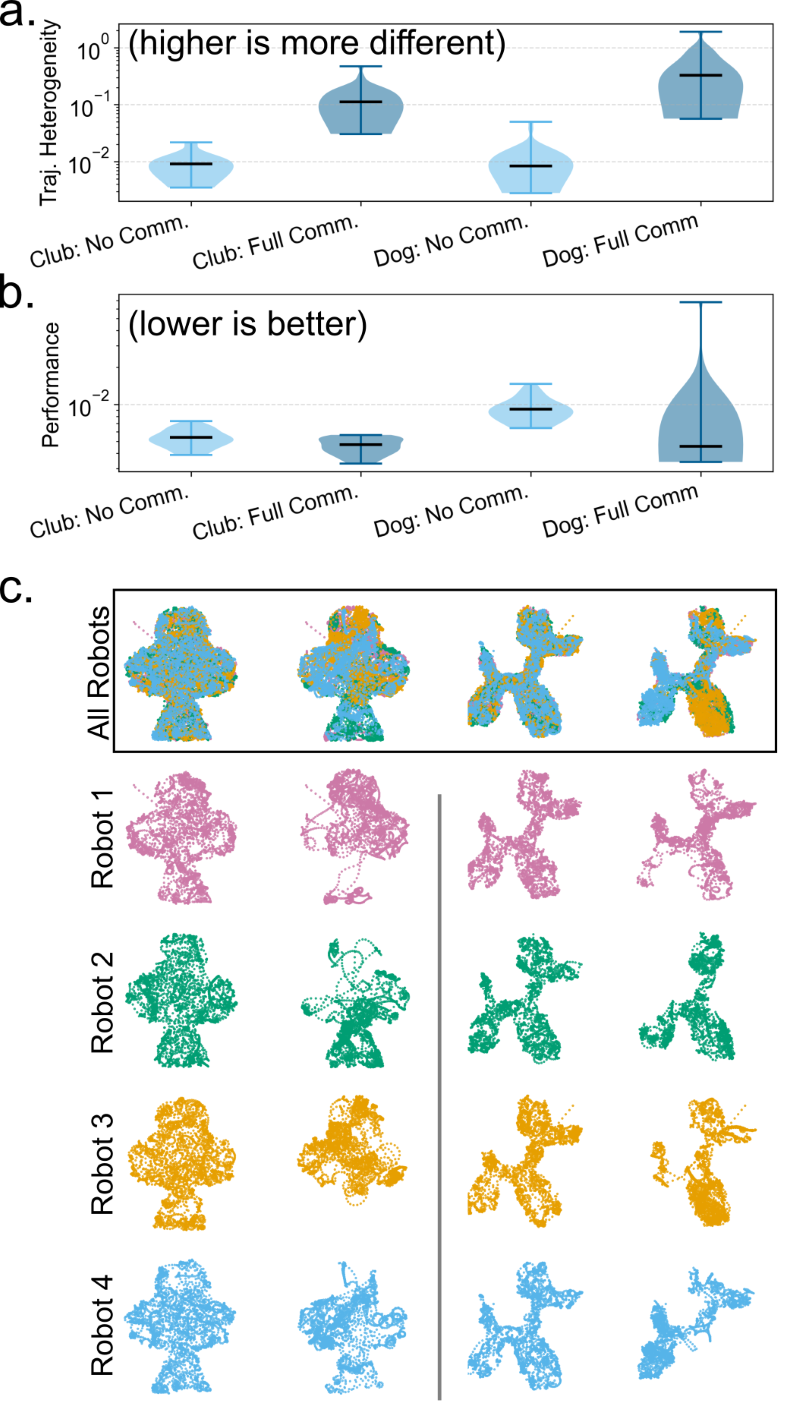}
    \caption{\textbf{Simulated results with and without communication.} Four robots, each running the ergodic control algorithm, were simulated for 25 trials. No communication means that each robot is separately running ergodic control. Full communication means that agents average their trajectory history to divide the task, using the decentralized ergodic control algorithm. The black lines in (a) and (b) are the median values. (a) Agent trajectories are more different from each other when they communicate. (b) Performance is the average ergodic metric over the trial. Agents perform similarly across all cases, but perform worse on the more complicated balloon dog objective. (c) For each objective, the no communication case is on the left and the full communication case is on the right. Individual agent trajectories are different when agents communicate and the same when they do not communicate.}
    \label{fig:sim_pic}
    % \vspace{-1.0em}
\end{figure}

\subsection{Assessing Trajectory Heterogeneity}
To measure task division, we use a metric on trajectory heterogeneity. To compute the trajectory heterogeneity between two agents, we compare the Fourier coefficients of individual agents using a norm outlined in Section~\ref{sec:ergodic_alg}. Instead of comparing a target distribution to agent trajectories, we compare the Fourier representation of the trajectories of agent $i$ and agent $j$. The trajectory heterogeneity metric between agent $i$ and agent $j$ is then:
\begin{equation} \label{eq:traj_comp}
    E_{i,j}(x_i(t), x_j(t)) = \,\sum_{k \in \mathbb{N}^n} \Lambda_k \left(c_{k_i} -c_{k_j} \right)^2
\end{equation}
where $\Lambda_k$ is a normalization coefficient. We compare the trajectory statistics of every agent compared to every other agent in a given trial. We then average the trajectory heterogeneity values to get an overall score per trial.

\subsection{Collision Handling}
Work in~\cite{Taylor2024} showed that agents using the ergodic control algorithm with shared states in their coverage domains are guaranteed to be co-located. As a result, agents will inevitably collide with each other during the patterning process. The presence of collisions creates a notable issue: for a system to be ergodic with respect to a set of states, coverage of all states of the system is necessary. To this end, we make a key assumption, shown in Equation~\ref{eq:collective_dynamics}, that agent velocities are uncorrelated. However, collisions mean that agents' velocities will be correlated, jeopardizing the ergodicity of agents' coverage. As our claims about the ergodicity of our system relate time-averaged behavior to coverage of states, correlations in agent behavior may bias their exploration and thus will compromise ergodicity with respect to the target distribution. 

% To explain why collisions affect the ergodicity of our system, we use the molecular chaos hypothesis: the assumption that the velocities of colliding particles are independent after collision and independent of the position of the collision~\cite{ehrenfest1990conceptual}. This assumption is commonly used in the kinetic theory of gases. By assuming the velocities of agents after a collision are uncorrelated, we can calculate the probability of two particles colliding by considering each particle separately. As our claims about the ergodicity of our system relate time-averaged behavior to coverage of states, correlations in agent behavior may bias their exploration and thus will not be ergodic with respect to the target distribution. 

% Consider two agents moving in a square domain with a uniform target distribution. Without collisions, each agent explores the space independently, and the combined trajectories approximate the uniform distribution as time progresses. Now, suppose the two agents collide while both are moving in the same direction. If they were to simply bounce elastically, their post-collision velocities would remain correlated: both agents would still tend to move in roughly the same direction. This would create local regions of over-sampling and other regions of under-sampling. These correlations accumulate over many collisions, and the time-averaged visitation frequency of states will deviate from the target distribution, violating ergodicity.

To resolve this issue, agents have a de-correlation protocol when they collide: each agent calculates a random direction in their respective half plane, which preserves their independence. Thus, each agent chooses a new velocity independently in a half-plane pointing away from their current heading. This ensures that post-collision trajectories are statistically independent. Then, the system recovers the property that long-term averages of the trajectories converge to the spatial distribution prescribed by the coverage objective.

\section{Simulations}
As discussed in Section~\ref{sec:robot_comm}, communication is achieved using the decentralized ergodic control algorithm in which agents share their trajectory history, average that history, and then synthesize control actions based on the collective progress~\cite{abraham2018decen}. Here, we simulate the system with two objectives with 25 trials for each communication style for four agent teams. The agent trajectories are shown in Figure~\ref{fig:sim_pic}. Each column represents a different communication style for the target images. The effect of communication is reflected visually in the agent trajectories, where task division increases with increasing communication. The median trajectory heterogeneity for communicating agents is higher for both objectives. Performance refers to the average value of the ergodic metric over a trial with respect to the agents' true dimple distribution. Lower values indicate better performance. Figure~\ref{fig:sim_pic}a and Figure~\ref{fig:sim_pic}b show the trajectory heterogeneity metric and task performance, respectively. Performance is similar across communication styles.
\begin{figure*}[t!]
\centering
\includegraphics[width=\textwidth]{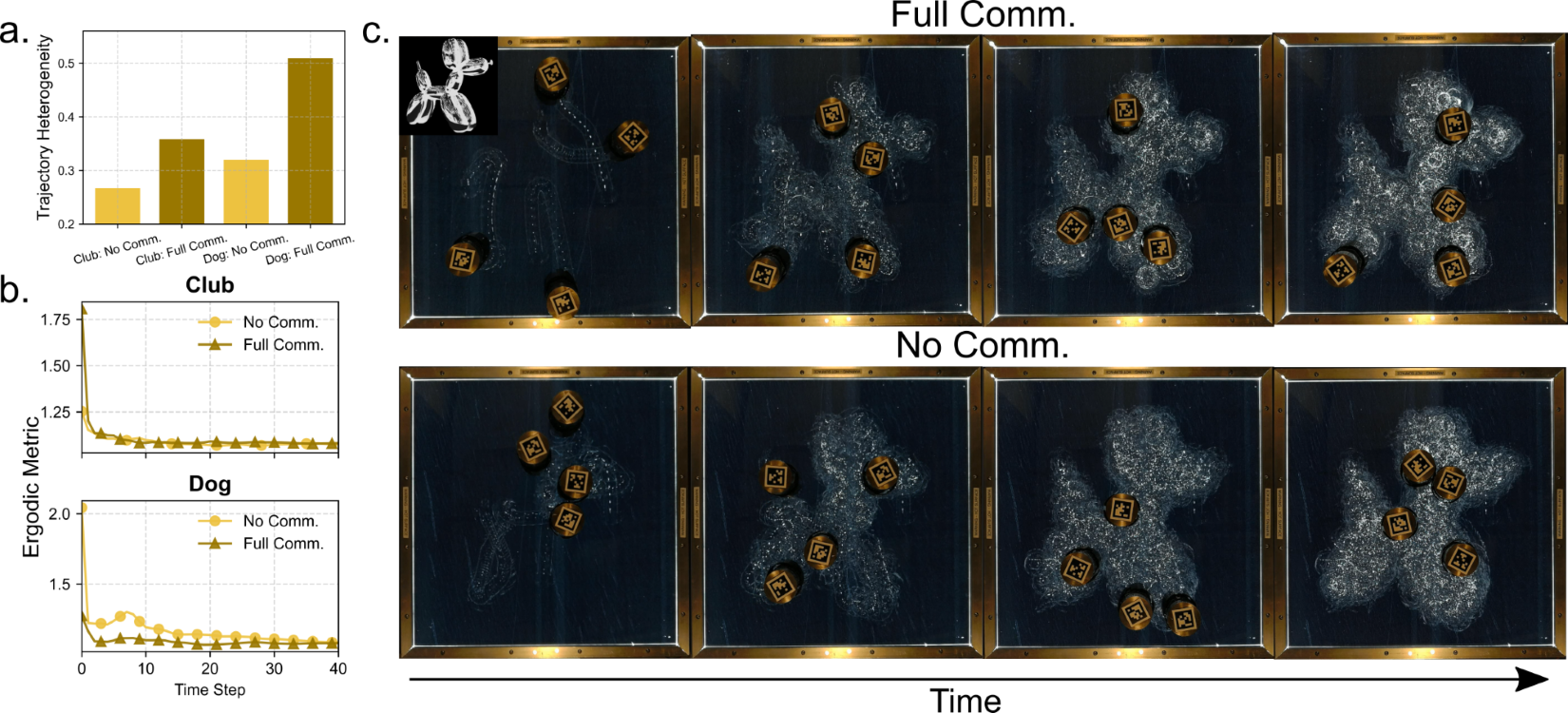}
    \caption{\textbf{Experimental results with and without communication.} Note that these results include one experimental trial for each case. (a) Trajectory heterogeneity score for each objective. Communicating agents have a higher trajectory heterogeneity score than non-communicating agents. The heterogeneity scores were higher for the dog objective, which is more complicated. (b) The system's ergodic metric over each trial. This shows a snapshot of the first 40 timesteps. The ergodic metric was minimized and stabilized to the shown values. Communication benefits the performance of the more complicated dog object, while it does not affect the performance of the club objective. (c) Images over time from the dog patterning experiment with 15 minutes total time. Elapsed time from left to right: 30 seconds, 2.5 minutes, 7.5 minutes, 15 minutes.}
    \label{fig:exp_pic}
\end{figure*}
\begin{figure}[t!]
\centering
\includegraphics[width=\columnwidth]{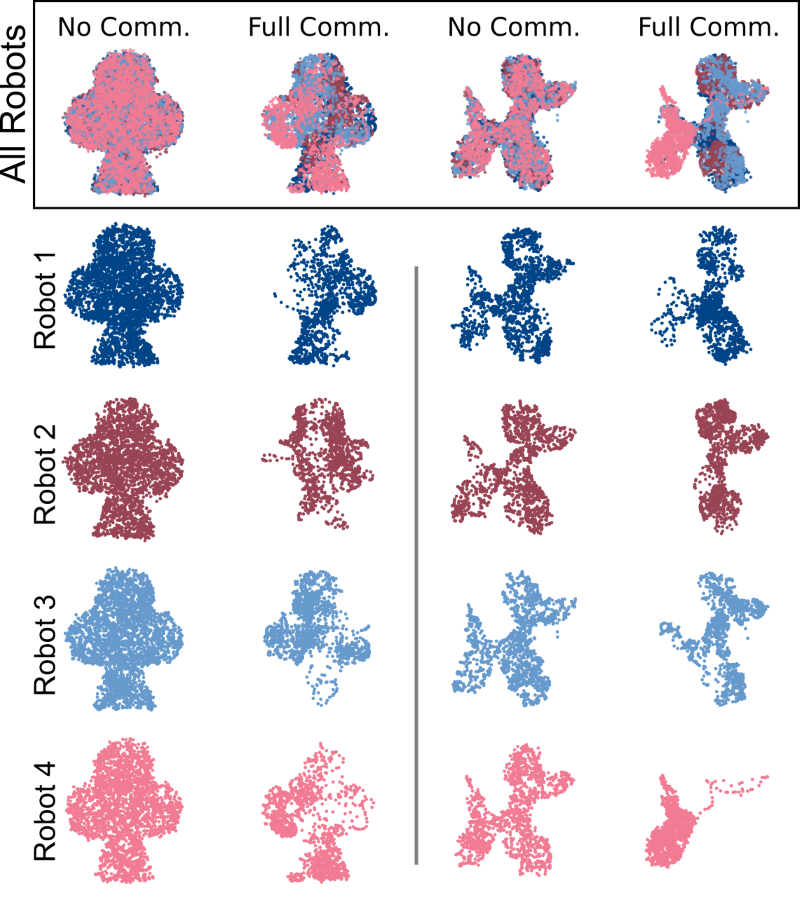}
    \caption{\textbf{Experimental trajectories for each robot.} Individual robot trajectories for each case. Robots that communicate can divide the task, taking responsibility for a part of the coverage area. Robots that do not communicate have redundant coverage, each covering the whole area.}
    \label{fig:exp_traj}
\end{figure}
\section{Experiments}
We performed experiments with four agents for each communication mode for the club and balloon dog objectives. For these experiments, the robots were tracked with an overhead camera and sent their true positions. As the robots were connected to a laptop through Bluetooth for localization, we communicated the robot's trajectory history to other robots through the same central hub. The robot still computes the ergodic control algorithm onboard, so this system can be described as distributed, but not fully decentralized.  In future work, we plan to have the robots localize themselves and communicate directly using Bluetooth. Rather than a metallic workpiece, the robots apply dimples to an acrylic sheet illuminated from its edges. Each dimple redirects the light toward the camera, making dimples easier to see from a distance than those on a metal surface.

The resulting patternings are shown in Figure~\ref{fig:exp_pic}. As in simulation, the communicating agents can divide the task better than the agents without communication. This can also be seen in the trajectory heterogeneity metric from Equation~\ref{eq:traj_comp}, also used with the simulated results. The experimental trajectory heterogeneity is shown in Figure~\ref{fig:exp_pic}a, and the individual agent trajectories are shown in Figure~\ref{fig:exp_traj}. No communication shows the least heterogeneity, while full communication shows the most differences between agent trajectories. We also compare the performance of the two communication cases, where, again, performance refers to the average value of the ergodic metric over a trial with respect to the agents' true dimple distribution. We see that for the club objective, both communication cases perform the same. For the dog objective, the no communication case performs worse than the full communication case at the beginning of the run. The dog objective is more complicated in both density specification and shape, making it possible for robots to become jammed in different areas, which can compromise performance.

\subsection{Functional Surface Verification}
Previous work has demonstrated that a patterning tool with the form factor used in our experiments can produce patterns that reduce friction due to hydrodynamic drag~\cite{landis2025patterning}. That work mounted the robot texturing tool in a computer-controlled motion stage, then rastered the tool over a workpiece at varying speeds and stepovers, producing surface patterns on small aluminum stubs. These stubs could then be evaluated in a rotary tribometer to determine their frictional performance. While this strategy showed that this tool could produce successful friction-reducing features and that density was a key control parameter, the patterns were not made by the robots themselves. This route was taken because the robots are not designed to produce exact pattern densities on workpieces much smaller than themselves, such as the stubs. Using a motion stage allowed sufficient control over pattern density to measure the effect of density relatively precisely while applying patterns to the small workpieces directly.

\begin{figure}[t]
\centering
\includegraphics[width=\columnwidth]{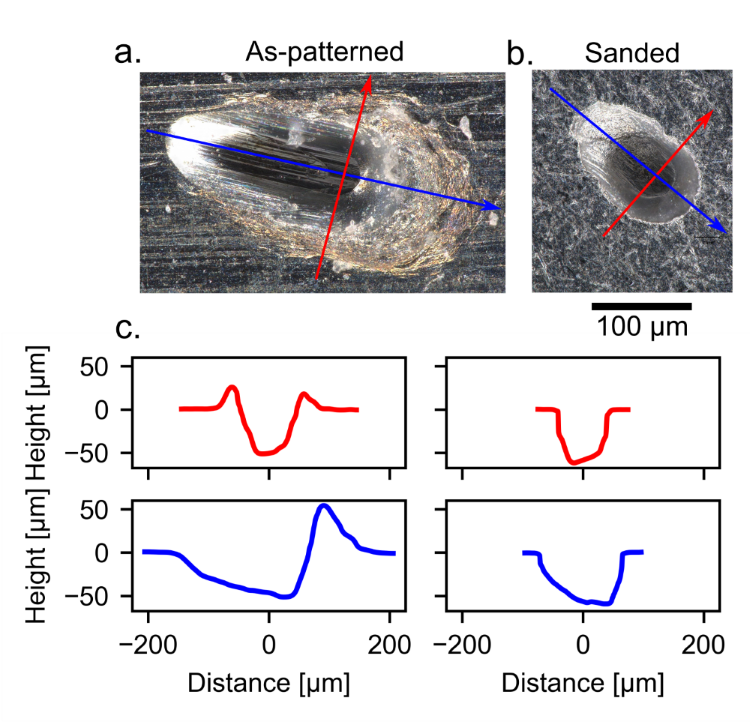}
    \caption{\textbf{Representative dimples on aluminum.} A typical dimple after patterning (a) and a similar dimple (b) after being sanded to remove raised edges. Depthmap slices (c) show the excess material removed by sanding as well as the asymmetry due to robot motion during patterning.}
    \label{fig:representative_dimples}
\end{figure}

Here, we show that the textures produced by the robots themselves have improved frictional performance, though we focus on the transition from the hydrodynamic to the mixed lubrication regime. A gradient target distribution of high to low density (from left to right) was loaded onto a single robot which patterned a 50 by 50~mm area on an aluminum 6061 sheet over 100 minutes, leaving approximately 20 thousand dimples on the surface. After patterning, the sample was sanded with 1500-grit sandpaper to remove excess material displaced by the indentation tool during dimpling (a standard procedure for Laser Surface Texturing~\cite{kovalchenko2005effect}). Microscope images of a typical dimple after patterning and a similar dimple after sanding are shown in Figure~\ref{fig:representative_dimples} along with depthmap information.

\begin{figure}[t!]
\centering
\includegraphics[width=\columnwidth]{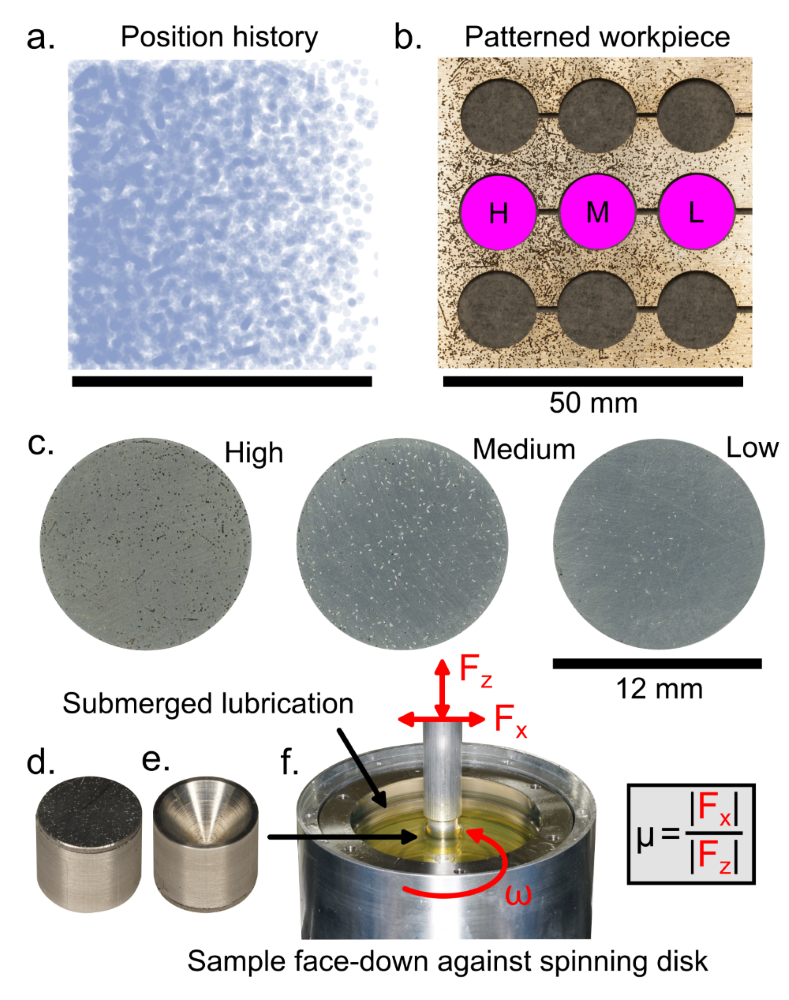}
    \caption{\textbf{Friction testing setup.} (a) Position history from robot and (b) resulting patterned workpiece with circular samples removed, (c) microscope images of samples taken from the high, medium, and low pattern density regions, (d) a sample bonded to its mount, (e) a reversed sample showing the leveling cone, and (f) the rotary tribometer setup showing the sample orientation, forces being measured, and input rotation.}
    \label{fig:friction_setup}
\end{figure}

\begin{table}[!]
    \centering
    \begin{tabular}{r|l}
        Duration [s] & Rotation Rates [rpm] \\
        30 & 400, 360, 320, 280, 240, 200 \\
        60 & 180, 160, 140, 120, 100, 90, 80, 70, 60, 50 \\
        90 & 45, 40, 35, 30, 25, 20 \\
        120 & 18, 16, 14, 12, 10, 8, 6, 4, 2
    \end{tabular}
    \caption{Rotary tribometer testing sequence.}
    \label{tab:sequence_settings}
\end{table}
To overcome the restriction on small samples, we pattern a large workpiece with our robots, then use Wire Electrical Discharge Machine (WEDM) to machine small circular samples, which are then glued to aluminum stubs, shown in Figure~\ref{fig:friction_setup}. The stubs have a cone machined to allow tip and tilt motion to ensure the sample face runs parallel to the disk of the rotary tribometer. The rotating disk of the tribometer is turned flat before each test. Three different texture densities were tested (high, medium, and low), corresponding to 7.5\%, 5.4\%, and 0.5\% area coverage (measured using ImageJ~\cite{schneider2012nih}), along with a blank, control sample.

Samples were submerged in 80W-90 gear oil (Lucas Oil) and positioned face-down, 23.874 mm from the center of disk rotation, giving 2.5 mm/s relative velocity per rpm. A series of discrete rotation rates from 2 to 400 rpm was applied to the disk, resulting in relative velocities from 0.005 to 1 m/s at the pattern-disk interface. The exact rotation rates and durations spent at each velocity are shown in Table~\ref{tab:sequence_settings}. We then applied a normal load and measured the resulting tangential load to calculate the friction coefficient over time. The test began by loading the sample to 5 N, accelerating to maximum velocity (i.e., 400 rpm), then increasing the load to 150 N. The relative velocity was then decreased in discrete steps over time. By starting at high relative velocity (i.e., within the hydrodynamic regime), the chance of wear damage affecting early results was minimized. 

\begin{figure}[t!]
\centering
\includegraphics[width=\columnwidth]{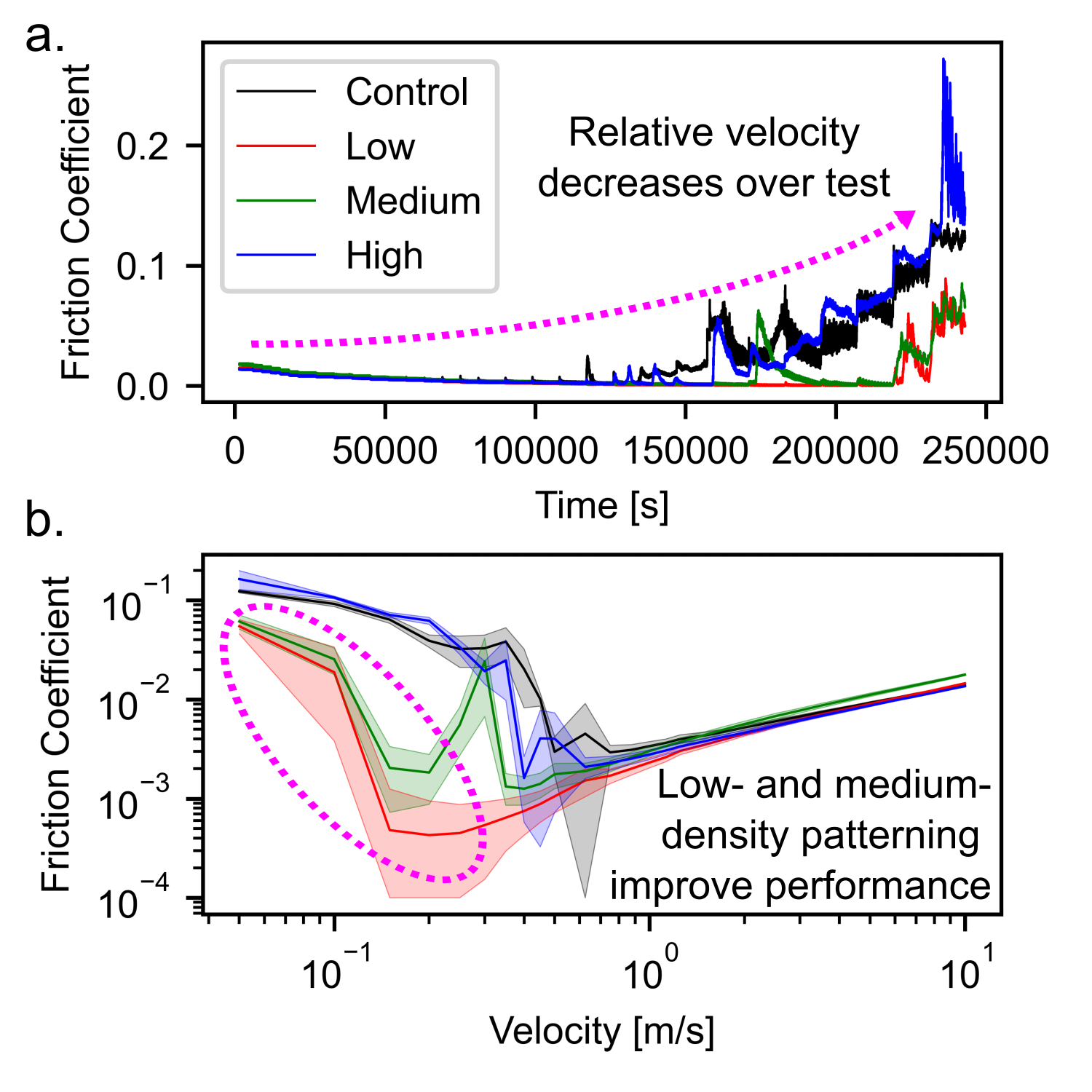}
    \caption{\textbf{Rotary tribometer results.} (a) Friction coefficient over test duration and (b) resulting Stribeck curves produced from the patterned workpiece. Low- and medium-density patterning reduces the onset of the mixed lubrication regime compared to a control, unpatterned surface. This result confirms the beneficial effect of robotic patterning for friction reduction.}
    \label{fig:friction_results}
\end{figure}

The results of friction testing are shown in Figure~\ref{fig:friction_results}. The raw data shows friction starting low, decreasing slightly over time (i.e., with decreasing velocity), then increasing toward the end of the test. The coefficient of friction for each discrete velocity step is averaged over time and plotted against relative velocity on a log-log plot, forming a Stribeck curve ~\cite{lu2006stribeck}. A Stribeck curve plots the friction coefficient against relative velocity, highlighting the boundary, mixed, and hydrodynamic lubrication regimes. The shaded region shows plus or minus one standard deviation. All samples show similar performance on the right side of the curve (i.e., in the hydrodynamic lubrication regime) but show differences as the velocity decreases (i.e., in the mixed and boundary regime). The low- and medium-density patterning appears to keep the friction coefficient low for longer, resulting in a two orders-of-magnitude difference between the control and low-density patterns at velocities around $2\times10^{-1}$. This verifies that the patterns produced by the robots have a beneficial effect that could be leveraged in real applications. 

\section{Conclusions and Future Work}

In this work, we demonstrated the feasibility of distributed multi-robot systems for scalable surface patterning. We showed that robots could collaborate on micro-patterning coverage tasks in simulated and hardware experiments. Further, we showed that robot-produced micro-patterns can impart beneficial physical properties on metallic surfaces. However, several avenues remain open for future research. 

First, a fully decentralized system will require that robots perform onboard localization, e.g., through SLAM, rather than relying on overhead vision. Developing such capabilities would enable true peer-to-peer coordination and remove the need for camera infrastructure in the environment. 

Second, many manufacturing processes involve sequential or complementary tasks, suggesting the potential for leader–follower interactions in which one group of robots generates patterns while others perform post-processing operations such as sanding or coating. Ensuring reliable, safe coordination in these settings motivates the use of control barrier functions beyond restricting the coverage domain. In future work, control barrier functions could capture both individual agent constraints and collective system-level requirements, such as mediating sequential tasks through onboard sensing of the state of the patterning task. 

Finally, the role of communication remains a critical design consideration. While sharing richer state information, such as trajectory history, enables coordination, using low-bandwidth communication techniques can maintain scalability. Future directions could include the development of adaptive or learned communication policies that determine what information should be exchanged. Communication may also include onboard sensing of the shared patterning task to track the progress of the patterning process. Together, these extensions would broaden the applicability of multi-robot patterning systems to more complicated, heterogeneous, and industrially relevant manufacturing settings.

% COMMENTED OUT BECAUSE OF DOUBLE BLIND REVIEW PROCESS
\section{Acknowledgments}
\label{sec:acknowledgements}
This research was supported by the National Science Foundation under Grant number CNS-2229170. Malachi Landis is supported by the National Defense Science and Engineering Graduate (NDSEG) Fellowship.
\balance 

\bibliographystyle{IEEEtran}
\bibliography{bib/iros2024, bib/decentralized_patterning, bib/active_learning}

\end{document}